# Implementation of the Principal Component Analysis onto High-Performance Computer Facilities for Hyperspectral Dimensionality Reduction: Results and Comparisons

Ernestina Martel [1,*], Raquel Lazcano [2], José López [1], Daniel Madroñal [2], Rubén Salvador [2], Sebastián López [1], Eduardo Juarez [2], Raúl Guerra [1], César Sanz [2] and Roberto Sarmiento [1]

[1] Institute for Applied Microelectronics (IUMA), University of Las Palmas de Gran Canaria (ULPGC), 35001 Las Palmas de Gran Canaria, Las Palmas, Spain; lopez@iuma.ulpgc.es (J.L.); seblopez@iuma.ulpgc.es (S.L.); rguerra@iuma.ulpgc.es (R.G.); roberto@iuma.ulpgc.es (R.S.)
[2] Universidad Politecnica de Madrid, 28031 Madrid, Spain; raquel.lazcano@upm.es (R.L.); daniel.madronal@upm.es (D.M.); ruben.salvador@upm.es (R.S.); eduardo.juarez@upm.es (E.J.); cesar.sanz@upm.es (C.S.)
* Correspondence: emartel@iuma.ulpgc.es; Tel.: +34-928-452876

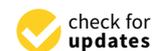

**Abstract:** Dimensionality reduction represents a critical preprocessing step in order to increase the efficiency and the performance of many hyperspectral imaging algorithms. However, dimensionality reduction algorithms, such as the Principal Component Analysis (PCA), suffer from their computationally demanding nature, becoming advisable for their implementation onto high-performance computer architectures for applications under strict latency constraints. This work presents the implementation of the PCA algorithm onto two different high-performance devices, namely, an NVIDIA Graphics Processing Unit (GPU) and a Kalray manycore, uncovering a highly valuable set of tips and tricks in order to take full advantage of the inherent parallelism of these high-performance computing platforms, and hence, reducing the time that is required to process a given hyperspectral image. Moreover, the achieved results obtained with different hyperspectral images have been compared with the ones that were obtained with a field programmable gate array (FPGA)-based implementation of the PCA algorithm that has been recently published, providing, for the first time in the literature, a comprehensive analysis in order to highlight the pros and cons of each option.

**Keywords:** hyperspectral imaging; dimensionality reduction; principal component analysis; Jacobi method; GPU; manycore; FPGA

## 1. Introduction

Hyperspectral imaging systems are nowadays considered as one of the most powerful remote sensing tools for acquiring precise information of the Earth's surface. These optical systems are able to provide images in which single pixels have information from across the electromagnetic spectrum of the scene under observation. In other words, whereas the human eyes see colors of visible light in mostly three bands (red, green, and blue), hyperspectral imaging divides the spectrum into many contiguous bands (typically hundreds), which may be even out of the visible part of the electromagnetic spectrum. The main advantage of this technique is that certain objects leave unique 'fingerprints' in the electromagnetic spectrum, which are known as spectral signatures. These 'fingerprints' enable the identification of the materials that make up a scanned object, providing a much better understanding of the scene under analysis [1].

However, the dimensionality of the space spanned by the pixels of a particular hyperspectral image is generally much lower than its number of spectral bands, which means that some bands of hyperspectral images usually provide redundant information that can be reduced by identifying an appropriate subspace. This allows for carrying on a spectral reduction of the captured hyperspectral data [2]. Dimensionality reduction algorithms have been extensively used in the literature because of two equally relevant main reasons. First, because it represents an essential processing step to reduce the huge amount of information of hyperspectral images. Effectively, raw data captured from current hyperspectral imaging sensors require, for their representation, a space whose number of dimensions equals the number of spectral bands provided, which is typically several hundreds. Such a value oversizes the minimum number of dimensions of the intrinsic supporting subspace. Therefore, to avoid the computational complexity that is derived from dealing with large dimensional spaces, the original hyperspectral data should be shrunk to keep the most relevant information. Second, because it has been demonstrated in the literature that such an isolation of the most significant information contained in the targeted image clearly benefits the results that are achieved by the ulterior hyperspectral image processing steps, such as unmixing, classification and target detection, just to name some [3–5].

Among the different techniques for decorrelating and reducing the amount of spectral information in hyperspectral images, the Principal Component Analysis (PCA) algorithm has become one of the most popular options [6]. PCA is based on projections techniques seeking for the best subspace for representing the collected hyperspectral data. Specifically, the PCA algorithm seeks the projections that best represent the hyperspectral data in terms of its variance. For this, it is necessary to calculate the eigenvalue decomposition of the covariance matrix, usually after centering the data in the average of each spectral observation. Then, by selecting only the eigenvectors that correspond to the largest eigenvalues, a reduction of the dimensionality of the original data is achieved while retaining a wealth of information (variance) in the data.

Although this PCA dimensionality reduction has proven to bring the aforementioned benefits to the whole hyperspectral image processing chain, it is also true that its utilization is not exempt from a formidable computational effort, which may compromise its use in time-sensitive (real-time or near real-time) applications. In these applications, the algorithms that are required for processing the data are typically implemented onto high-performance computing architectures in which the operations involved are executed in parallel devices [7–9]. In this sense, different research groups have published works that deal with the implementation of the PCA algorithm onto different high-performance computer architectures, such as Field Programmable Gate Arrays [10], graphic processing units (GPUs) [11], and cloud computing infrastructures [12], which represent the most recent efforts in the literature toward this goal. Unfortunately, none of these works provides comparisons between the different high-performance computing devices, and their benefits and weaknesses, according to the characteristics of the different applications remain unknown for the reader. To the best of the authors' knowledge, this work represents the first step towards this goal, in the sense that it compares the performance that is achieved by different high-performance computing platforms when applying the PCA algorithm to a set of hyperspectral images. Precisely, this work is focused on two main aspects. First, on efficiently implementing, by means of the Jacobi technique, the PCA algorithm onto two popular high-performance computing platforms, such as a GPU from NVIDIA and a manycore from Kalray, which represents the main innovation that is introduced in this work. Second, on providing a comprehensive comparison of the suitability of the implementation of the Jacobi algorithm on the previously mentioned platforms and a third one, collected form the state-of-the-art, which represents the most recent work in the literature in which the PCA algorithm, also through the Jacobi technique, is implemented onto an FPGA [10]. Last but not least, we would like to recognize the existence in the state-of-the-art of other hyperspectral dimensionality reduction techniques, such as the ones published in [13–16], which have demonstrated to outperform PCA in terms of dimensionality reduction and computational complexity capabilities. However, we have decided to focus our efforts in the implementation of the PCA algorithm as nowadays represent the *de facto* technique that is utilized in many hyperspectral image processing chains so far published.

The rest of the paper is organized as follows. Section 2 reviews the PCA algorithm and the Jacobi method. Section 3 introduces the main characteristics of the high-performance computing devices that are considered in this work, i.e., GPUs and manycores, while Sections 4 and 5 give a detailed explanation about how the Jacobi-based PCA algorithm has been efficiently implemented onto an NVIDIA GPU and a Kalray manycore, respectively. Finally, Section 6 presents, analyzes, and compares the results that were obtained with both devices and with the FPGA-based implementation published in [10], while Section 7 draws the main conclusions of this work.

## 2. Dimensionality Reduction by Means of the PCA Algorithm

Principal Component Analysis (PCA) is a popular method that is used to spectrally compact high dimensional datasets [2]. In order to capture most of the variation of the targeted dataset, PCA projects the data onto an orthogonal subspace with a smaller dimension, which represents a very efficient technique to eliminate the existing redundancies among adjacent highly correlated bands. With PCA, an optimal dimension subspace in which the data volume lies can be discovered. PCA is optimal in the sense that it finds a number of orthogonal directions, smaller than the data space dimension, that, once the data volume is projected onto them, maximizes the total variance of the projections, i.e., maximizes the total variance from the original image, which is retained within the built-in subspace. Furthermore, the variance of the different projections is rearranged as a decreasing function of the sequence of selected orthogonal directions.

PCA consists of four main stages. The pseudocode is provided in Algorithm 1, together with an approximation of the number of Floating Point Operations (FLOPs) that each computing step involves. First, the data volume is moved to be recentered around the reference origin point. This data preprocessing is achieved by computing and removing the average value of each spectral band (see line 2 of Algorithm 1). Secondly, the covariance matrix of the data volume is computed as the product of the preprocessed data matrix and its transpose (line 3). Next, the eigenvectors that are associated to the covariance matrix are extracted (line 4). Finally, every pixel of the original image is projected onto a subset of eigenvectors to achieve the dimensionality reduction (lines 5 and 6).

| Algorithm 1: Principal Component Analysis | Number of FLOPS |
|---|---|
| 1  **Input:** Hyperspectral image Y (N × M matrix), N pixels, M bands | |
| 2  X = BandAverageRemoval (Y) | $2(N \times M) + M$ |
| 3  Covariance matrix, C = $X^T \cdot X$ | $2 \times N^2 \times M^2$ |
| 4  E = EigenvectorDecomposition (C), e eigenvectors computed | $4 \times M^3$ |
| 5  Projection Matrix, Q = Y·E | $e \times N(2 \times M - 1)$ |
| 6  Q' = MatrixColumnRemoval (Q, p), p principal components | |
| 7  **Output:** Reduced hyperspectral image Q' (N × p matrix) | |

According to the rightmost column in Algorithm 1, the number of FLOPS required for the dimensionality reduction of a hyperspectral image composed by 512 × 512 pixels (N = 262,144) and 224 bands (M = 224) can reach the total amount of more than $10^{15}$ FLOPS, which reinforces the crucial role that is played by high-performance computing facilities for applications under real-time constraints.

As stated in Algorithm 1, one of the most computationally demanding processes within the PCA algorithm is the eigenvector decomposition of the covariance matrix. A naive method to find an eigenvector decomposition of a matrix consists of finding the roots of the characteristic polynomial of the matrix, and then, solving a set of systems of linear equations using, for instance, Gaussian elimination [17]. However, this procedure is not feasible for matrices as large as those that are involved in hyperspectral image processing.

Iterative algorithms have been extensively employed to compute eigenvalues and eigenvectors. This class of algorithms refines an approximation of the eigenvectors until a convergence criterion is satisfied. Different iterative methods have been described in the literature [17]. Among them, the Jacobi method [18–20] is distinguished. In this method, the sequential nature of the iterative process is balanced with the possibility to exploit its inherent parallelism. In addition, the method extracts all of the eigenvectors of the matrix, a relevant feature for the computation of the PCA algorithm. Its main objective is to approximate the input image to a diagonal matrix by iteratively zeroing all of the off-diagonal elements thanks to the application of successive planar rotations, known as Jacobi rotations. It should be noted that this methodology only applies to real and symmetric matrices. To better understand this method, a toy example with an input matrix of $2 \times 2$ is provided.

Given a real and symmetric matrix $A = \begin{pmatrix} a & b \\ b & c \end{pmatrix}$, and a rotation matrix $P = \begin{pmatrix} \cos \alpha & \sin \alpha \\ -\sin \alpha & \cos \alpha \end{pmatrix}$, the Jacobi method finds a value of $\alpha$ such that the operation shown in (1) results in a diagonal matrix, $B$.

$$B = P^{-1} \cdot A \cdot P \tag{1}$$

By substituting $A$ and $P$ in Equation (1) and operating, the values of $\cos \alpha$ and $\sin \alpha$ can be deduced, obtaining the expressions (2) to (5). After applying these expressions, the element $b$ in matrix $A$ would be zeroed, thus converting the original matrix in a diagonal one.

$$m = \frac{2 \cdot b}{c - a} \tag{2}$$

$$t = \frac{-1 + \sqrt{1 + m^2}}{m} \tag{3}$$

$$\cos \alpha = \frac{1}{\sqrt{1 + t^2}} \tag{4}$$

$$\sin \alpha = t \cdot \cos \alpha \tag{5}$$

Extending this method to matrices of greater dimension is just a matter of generalizing these expressions. As can be inferred from (1), the rotation matrix $P$ must present the same dimensions as the input matrix, which in the case under study is the covariance matrix of the original hyperspectral image, $C$. Hence, its structure is similar to that shown in (6). As it can be inferred from (6), $P$ is similar to an identity matrix, except for the $P_{ii}$, $P_{ij}$, $P_{ji}$, and $P_{jj}$ values.

$$P = \begin{pmatrix} 1 & \cdots & & \cdots & 0 \\ \vdots & \cos \alpha & & \sin \alpha & \vdots \\ \vdots & -\sin \alpha & & \cos \alpha & \vdots \\ 0 & \cdots & & \cdots & 1 \end{pmatrix} \tag{6}$$

Furthermore, as of now, there are more off-diagonal elements, $\alpha$ should depend on the element selected to be zeroed, $C_{ij}$, where $i$ represents the rows of the covariance matrix and $j$ represents the columns. As a result, expression (2) is generalized as shown in (7). On the contrary, expressions (3) to (5) remain unchanged.

$$m = \frac{2 \cdot C_{ij}}{C_{jj} - C_{ii}} \tag{7}$$

This process nulls one element of the matrix at a time, so in each iteration ($k$) the operation shown in (8), which is a generalization of that shown in (1), must be performed. Specifically, each iteration recalculates the rotation matrix and generates $C_k$, where the element $C_{ij}$ is now zeroed, as well as its symmetric counterpart. Then, the next iteration will consider this matrix as its input.

$$C_k = P_k^T \cdot C \cdot P_k \qquad (8)$$

Although it can be deduced that each iteration can undo some of the zeros that were previously achieved, it has been demonstrated that the magnitude of all off-diagonal elements is reduced after each iteration. As a result, these operations are iteratively repeated until a provided stop condition, $\varepsilon$, is reached, that is, until all of the off-diagonal elements are smaller than this stop factor. Equation (9) summarizes the complete Jacobi method, where $K$ represents the total amount of iterations, $C_K$ is a diagonal matrix containing the eigenvalues of the input image, and $P_i$ are the successive Jacobi rotation matrices. Similarly, the associated eigenvectors can easily be computed, as shown in (10), where the eigenvectors are stored in a column-wise order.

$$C_K = P_k^T \cdot \cdots \cdot P_1^T \cdot C \cdot P_1 \cdot \cdots \cdot P_k \quad 1 \leq k \leq K \qquad (9)$$

$$E = P_1 \cdot P_2 \cdot P_3 \cdot \cdots \cdot P_K \qquad (10)$$

The convergence of Jacobi method has been successfully proven for two different strategies [18], depending on the order in which the elements are selected to be zeroed. The classical method selects the largest off-diagonal element in each rotation, whereas the cyclic method chooses the elements in a given order, e.g., row by row or column by column. Although the former minimizes the number of iterations, the latter is faster, since it bypasses the location of the largest element in each iteration, which is a quadratic-order operation. For that reason, the latter is the method that is selected in this study.

The main advantage of the Jacobi method is that it is inherently parallelizable. After the computation of each iteration, matrix $C_k$ matches with $C_{k-1}$, except in rows and columns $i$ and $j$. As a result, several rotations can be computed simultaneously. Specifically, two elements can be zeroed in parallel as long as they do not share any position, i.e., they are not located in the same row or column. However, the computation of all the eigenvectors in large matrices like the ones that are involved in hyperspectral imaging applications is still a computationally intensive problem, although it can benefit from parallelization techniques, like the Jacobi method. Hence, for time-sensitive applications, it is advisable to use high-performance computing platforms in which the whole process is effectively accelerated by means of their internal parallel computing structure.

## 3. High Performance Computing in Graphics Processing Units and Manycores

Hyperspectral imaging applications under timing constraints require to rapidly process a huge amount of data. In these applications, the algorithms used for processing the data, like the dimensionality reduction step targeted in this paper, are typically implemented onto high-performance computing platforms in which the operations that are involved can be executed in parallel as far as these algorithms exhibit an inherent parallelism. Among the different high-performance computing devices that are currently available in the market, graphic processing units (GPUs) and manycores have gained popularity since their multiple processors are easier to program than field programmable gate arrays (FPGAs) and dedicated hardware circuits.

In the past, one way to improve processor performance was to increase the processor clock rate. However, this technique generated excessive power dissipation. A suitable solution is to increase the number of cores in the processor, and consequently, to provide support for a larger number of threads. In this context, a single-chip multiprocessor is currently known as a multi-core processor. More specifically, a manycore processor is a type of multi-core processor containing a large number of independent cores, typically several hundreds, which are designed to achieve the execution of a large number of tasks in parallel. Key factors that impact the performance of manycore processors are the internal core communication methods and the mechanisms to interact with the world outside of the processor. It is worth noting that, besides the number of cores, multi-core processors are differentiated from manycores in the number of resources devoted to figure out the implicit parallelism available in a single thread.

GPUs provide much larger availability of processing cores than standard CPU processing, with smaller processing capability of the cores and small control units that are associated to each core (see Figure 1). Hence, GPUs are appropriate for algorithms that need to execute many repetitive tasks with fine grain parallelism and little coordination between tasks.

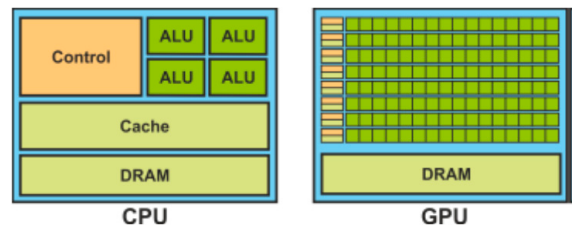

**Figure 1.** CPU vs Graphics Processing Unit (GPU) architecture.

Figure 2 shows the architecture of the GPU, which can be seen as a set of multiprocessors. Each multiprocessor is characterized by a single instruction multiple data (SIMD) architecture, i.e., in each clock cycle each processor of the multiprocessor executes the same instruction, but operating on multiple data streams. Each processor has access to a local shared memory and also to local cache memories in the multiprocessor, while the multiprocessors have access to the global GPU (device) memory.

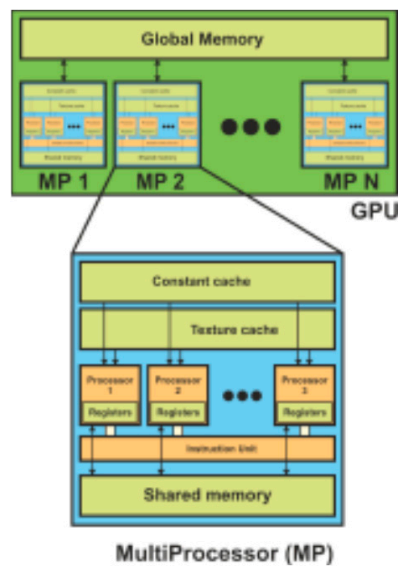

**Figure 2.** General hardware architecture of a NVIDIA GPU.

Within manycore processors, a typical architecture consists of a hierarchical structure that groups $M$ processing elements in $N$ clusters, which in turn, communicate with the external world through an Input/Output (I/O) subsystem. In addition, a Network-on-Chip (NoC) is usually employed for cluster interconnection. A clear example of this architecture is described in [21]. The Kalray MPPA-256-N (where MPPA stands for *Massively Parallel Processor Array*) is a single-chip manycore processor that targets low to medium volume applications with requirements such as low energy per operation and time predictability. This architecture is a homogeneous Multiprocessor System-on-Chip (MpSoC) mainly composed of 256 Very Long Instruction Word (VLIW) Processing Elements (PEs) distributed over 16 clusters and running at 400 MHz—see Figure 3. All of the clusters are interconnected through two NoCs, one for control information (C-NoC) and another for data communications (D-NoC) [22]. Likewise, external communications are handled with two I/O subsystems. Each of them contains a quad-core processor with a shared D-cache, four Direct Memory Accesses (DMAs), on-chip memory and a Double Data Rate (DDR) controller to access up to 64GB of external Double Data Rate (DDR) memory banks. Additionally, they run either a rich Operating System (OS) as Linux or a Real-Time Operating System (RTOS).

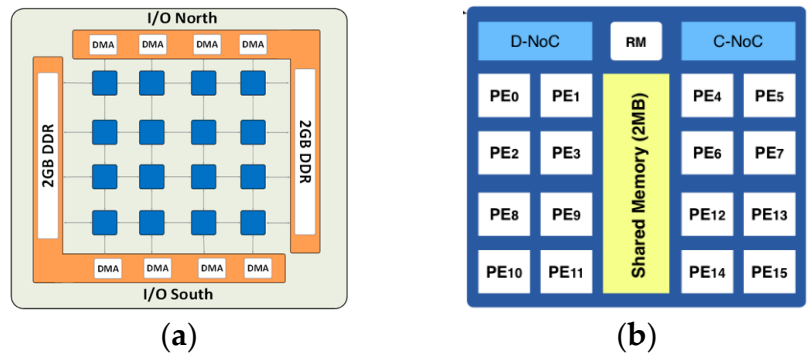

**Figure 3.** The Kalray Massively Parallel Processor Array (MPPA)-256-N architecture: (**a**) chip and (**b**) cluster.

The clusters are the basic processing unit of the MPPA architecture. Besides the 16 PEs, they contain one Resource Management (RM) core responsible for running NodeOS operating system and handling, events, and interruptions. They also contain 2 MB of shared memory organized in 16 parallel banks of 128 KB each. This shared memory provides a bandwidth of 38.4 GB/s, providing a fair trade-off between power, area, latency, and bandwidth [21], although it has been demonstrated to be the main limitation in data-intensive applications [23]. Each cluster also gathers a DMA engine for handling the communications between the NoC and the shared memory.

Finally, as described in [21], each PE or RM core implements a five-way VLIW architecture, including a multiply and accumulate floating-point unit, two arithmetic and logic units, together with a load/store unit and a branch and control unit, which allows up to 800 MFLOPS on 32-bit data at 400 MHz. These five execution units are interconnected through 64 32-bit general-purpose registers (GPRs). Furthermore, each PE also includes a Memory Management Unit (MMU) for virtual memory support and process isolation.

## 4. GPU Implementation of PCA Jacobi

This section focuses on the parallel implementation of the PCA Jacobi algorithm onto NVIDIA GPUs. For this purpose, we first present some Compute Unified Device Architecture (CUDA) preliminary issues and then we detail the implementation itself.

*4.1. Preliminary Issues*

Programs can be accelerated by moving the computationally intensive portions of the code onto a GPU. NVIDIA CUDA (*Compute Unified Device Architecture*) is a parallel computing platform and model that enables improvements in computing performance by harnessing the power of NVIDIA GPUs. CUDA programming model differentiates the CPU and its memory as the host and the GPU and its memory as the device. In the GPU, algorithms are constructed by chaining the so-called kernels, which define the minimum units of computation that are performed in the devices. Thereby, data-level parallelism is exposed to hardware, and kernels can be applied concurrently without any sort of synchronization. The kernels perform a kind of batch processing arranged in the form of a grid of blocks, as displayed in Figure 4, where each block is composed by a group of threads that share data efficiently through the shared local memory and synchronize their execution for coordinating accesses to memory. There is a maximum number of threads that a block can contain, but the number of threads that can be concurrently executed is much larger (several blocks executed by the same kernel can be managed concurrently, at the expense of reducing the cooperation between threads since the threads in different blocks of the same grid cannot synchronize with the other threads).

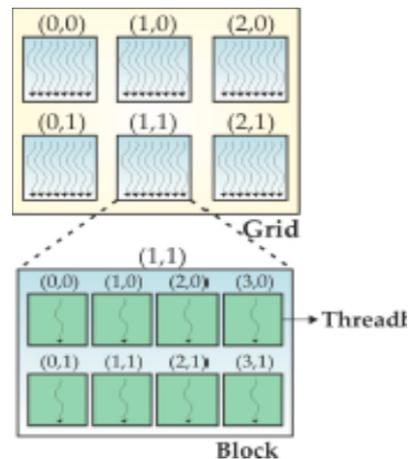

**Figure 4.** Processing kernels in the GPU: grids of blocks with computing threads.

The implementation of the PCA Jacobi algorithm that is described in Algorithm 1 was developed using CUDA onto NVIDIA proprietary GPU, and both, the cuBLAS and the Thrust library. The cuBLAS library [24] is an implementation of BLAS (*Basic Linear Algebra Subprogram*) on top of the NVIDIA CUDA runtime. cuBLAS provides functions whose execution takes place on the device, exploiting the maximum degree of parallelism. Moreover, Thrust [25] is a C++ template library for CUDA based on the Standard Template Library (STL). Thrust allows for implementing high performance parallel applications with minimal programming effort through a high-level interface that is fully interoperable with CUDA C. Moreover, Thrust provides host and device vector containers that release CUDA programmers from allocating memory, and transferring data between the host and device memory. Furthermore, Thrust allows for exploiting C++ iterators along with vector containers. In the case of vector containers, which are really just arrays, iterators can be thought as pointers to array elements. When a Thrust function is called, it inspects the type of the iterator to determine whether to use a host or a device implementation. This process is known as static dispatching, since the host/device dispatch is resolved at compile time. This implies that there is no runtime overhead to the dispatch process. Last but not least, Thrust provides a rich collection of data parallel primitives, such as scan, sort, and reduce, which can be composed together to implement complex algorithms with concise and readable source code. By describing computation in terms of these high-level abstractions, Thrust is provided with the freedom to select the most efficient implementation automatically. As a result, Thrust can be utilized in rapid prototyping of CUDA applications, where programmer productivity matters most, as well as in production, where robustness and absolute performance are crucial.

*4.2. CUDA Implementation of PCA Jacobi*

This section explains the CUDA implementation of the PCA Jacobi Algorithm that is presented in Algorithm 1, following the four stages of PCA Jacobi Algorithm 1: Image preprocessing; Covariance computation; Eigenvector decomposition; and, Projection and reduction. Apart from these stages, we first focus on the data initialization.

1. *Data Initialization*. The CUDA implementation of the PCA Jacobi algorithm requires the transfer of the hyperspectral image from the CPU (host) memory to the global memory of the GPU (device). Because some operations of the PCA Jacobi algorithm are implemented by means of cuBLAS library, which uses column-major storage, the hyperspectral image in column-major order has been transferred from the host to the GPU memory. Moreover, in order to simplify the transfer operation, the Thrust vector containers have been used for both, host and device. With this approach, we could completely avoid the need to manually manage the memory on the host and device using a Thrust vector for storing our data.

   In Code 4.1 the hyperspectral image is loaded into a Thrust host vector, *h_image_in*, by means of an iterator (lines 1.1, 1.2, 1.3, 1.4): in line 1.1, the filename with the hyperspectral image is passed as an argument of the command line; in lines 1.2, 1.3, and 1.4, the original image is loaded into a Thrust host vector container with *BANDS* × *PIXELS* elements, *h_image_in*. Finally, this image is transferred to the device memory, *d_image_in*, by means of *thrust::copy* (lines 1.5, 1.6). The result is that the hyperspectral image is loaded into the device as a sequence of bands. This is, the image matrix is row-major ordered. This fact must be considered whenever a cuBLAS function is called because cuBLAS uses column-major storage. This issue is achieved by using the transpose parameter whenever a cuBLAS function is called with the hyperspectral image.

   *Code 4.1. Data initialization*

   ```
   1.1    ifstream ifile(argv [1]);
   1.2    thrust::host_vector<float> h_image_in(BANDS * PIXELS);
   1.3    istream_iterator<float> beg(ifile), end;
   1.4    thrust::copy(beg, end, h_image_in.begin());
   1.5    thrust::device_vector<float> d_image_in(BANDS * PIXELS);
   1.6    thrust::copy(h_image_in.begin(), h_image_in.end(), d_image_in.begin());
   ```

2. *Image preprocessing*. In this stage, the original hyperspectral image is going to be mean-centered. In order to parallelize this step, we have divided it into two sequential calculations: the average per every band of the hyperspectral image and the subtraction of this average from the original hyperspectral image.

   2.1. *Average per band calculation*. For each band of the original image (*BANDS* × *PIXELS* matrix), its average is calculated and stored into a *BANDS* array. In order to calculate the average per band in parallel with the cuBLAS library, the *BANDS* × *PIXELS* matrix is multiplied with ones vector (vector with *PIXELS* elements) by means of the *cublasSgemv* function. According to the cuBLAS documentation, *cublasSgemv* performs the following matrix-vector multiplication:

   $$y = \alpha \, op\,(A)\,x + \beta \, y \qquad (11)$$

   where *A* is a $m \times n$ matrix that is stored in column-major format located in the GPU (device), *x* and *y* are vectors, which must be also located in the GPU, and *α* and *β* are scalars. The operand *op( . . . )* denotes the use or not of the transpose operation. In our average calculation:

   - *A* is the *BANDS*×*PIXELS* matrix which contains the original hyperspectral image in column-major format. According to Code 4.1, this matrix is already in the device, *d_image_in* (see lines 1.5 and 1.6 in Code 4.1). Moreover, because this matrix is in row-major format, it needs to be transposed in the cuBLAS function.
   - *x* is a vector with *PIXELS* elements filled to one. This vector is called *d_ones* and it has been filled to 1 by means of Thrust (see line 2.1.2 of Code 4.2.1).
   - *α* is set to 1/ *PIXELS*, and *β* is set to 0 (line 2.1.3 of Code 4.2.1).

- *y* is the array with *BANDS* elements that stores the average per band. In Code 4.2.1 this vector is called *d_means* and it is filled with the result of *cublasSgemv* function (see lines 2.1.1 and the next to last parameter of cuBLAS function in line 2.1.4).

---

*Code 4.2.1. Average per band calculation*

---

```
2.1.1    thrust::device_vector<float> d_means(BANDS); // to store average per band
2.1.2    thrust::device_vector<float> d_ones(PIXELS, 1.f);   //ones vector
2.1.3    float alpha = 1.f / (float) PIXELS; float beta = 0.f;
2.1.4    cublasSgemv(handle,              //handle for use cuBLAS
                    CUBLAS_OP_T,          //the d_image_in is transposed
                    PIXELS, BANDS,        //rows and columns of d_image_in
                    &alpha, thrust::raw_pointer_cast [1](d_image_in.data()),
                    PIXELS,               //leading dimension of d_image_in
                    thrust::raw_pointer_cast(d_ones.data()),
                    1,                    //leading dimension of d_ones
                    &beta,                //β scalar
                    thrust::raw_pointer_cast(d_means.data()),
                    1 );                  //leading dimension of d_means array
```

---

[1] `thrust::raw_pointer_cast` converts the device vector container to a raw pointer.

2.2. *Subtraction of the average per band from the original hyperspectral image.* For each band of the original image, the average of this band must be removed. The subtraction of the average of a band from every pixel in such band is carried out by means of a Thrust transformation. Transformations are algorithms that apply an operation to each element in a set of (zero or more) input ranges and then store the result in a destination range. The transformation *thrust::transform* in line 2.2.2 of Code 4.2.2, uses the function object *thrust::minus* to subtract a device vector, *d_image_in*, from another device vector, *d_means*. As *d_means* is a vector with *BANDS* elements and *d_image_in* contains *PIXELS* columns with *BANDS* elements, the *d_means* array must be subtracted *PIXEL* times for all of the pixels in the original hyperspectral image, *d_image_in*. This purpose is achieved by combining three Thrust fancy iterators: *permutation_iterator*, *transform_iterator* and *counting_iterator*. Each of these iterators is created by a function called *make_XXX*, where *XXX* denotes the name of the Thrust fancy iterator. The *permutation_iterator* permutes the range of values in *d_means* by the index scheme obtained by the *transform_iterator*. In general, *permutation_iterator* allows for operating on a specific set of values in a sequence instead of the entire sequence. The *transform_iterator* changes the range generated by the *counting_iterator* (linear index beginning in 0) to its associated row index by calling the *linear_index_to_row_index* functor (A C++ *functor* is an object that acts just like a function, but it has the advantage of being stateful, meaning that it can keep data reflecting its state between calls) (see line 2.2.1 in Code 4.2.2). This last functor transforms the linear index that is generated by the *counting_iterator* to the row index of a matrix.

*Code 4.2.2. Subtraction the average per band from the original hyperspectral image*

```
2.2.1   template<typename T>
        struct linear_index_to_row_index: public thrust::unary_function<T, T> {
            T Ncols; // --- Number of~columns
            __host__ __device__ linear_index_to_row_index(T Ncols):
                Ncols(Ncols) {
        }
            __host__ __device__ T operator()(T i) {
              return i / Ncols;}};

2.2.2   thrust::transform(d_image_in.begin(), d_image_in.end(),
        thrust::make_permutation_iterator
        (d_means.begin(),
        thrust::make_transform_iterator(
        thrust::make_counting_iterator(0),
        linear_index_to_row_index<int>(PIXELS))),
        d_image_in.begin(),thrust::minus<float>());
```

3. *Covariance computation*. This stage computes the covariance matrix associated with the original image in parallel in the GPU (device) by means of Thrust and the cuBLAS library. The process to calculate this covariance matrix is carried out by the *cublasSgemm* function, which multiplies the preprocessed matrix by its transpose. According to the cuBLAS documentation, *cublasSgemm* performs the matrix-matrix multiplication, following the next operation:

$$C = \alpha \, op(A) \, op(B) + \beta \, C \tag{12}$$

where $A$, $B$, and $C$ are matrices in column-major format located in the GPU (device), and $\alpha$ and $\beta$ are scalars. $A$ is an $m \times k$ matrix, $B$ is a $k \times n$ matrix, and $C$ is an $m \times n$ matrix. As mentioned before, the *op(...)* denotes the use or not of the transpose operation. In our covariance matrix calculation:

- $A$ and $B$ are *BANDS* × *PIXELS* matrices that contain the preprocessed hyperspectral image. According to our *Data Initialization code*, these matrices are already in the device as *d_image_in* (see lines 1.5 and 1.6 of *Data Initialization code*). Moreover, because *d_image_in* is in row-major format, it needs to be transposed in cuBLAS function. This is the reason why *op(A)* must be *CUBLAS_OP_T* (transposed *d_image_in*) and *op(B)* must be *CUBLAS_OP_N* (original *d_image_in*, not transposed) (see call to *cublasSgemm* function in line 3.3).
- $m$ is equivalent to *BANDS*, $n$ to *BANDS* and $k$ to *PIXELS*.
- $\alpha$ is set to 1, and $\beta$ is set to 0 (line 3.2 of the *Covariance computation code*).
- $C$ is the result matrix that corresponds with the covariance matrix, *d_cov*, a *BANDS* × *BANDS* matrix (see lines 3.1 and the next to last parameter in the cuBLAS function).

The last step for the covariance computation stage is to normalize the result of *cublasSgemm* using a Thrust transformation along with a Thrust iterator, as it is shown in line 3.4 of code 4.3. In this normalization, the elements of the *d_cov* device vector are divided by *PIXELS-1*.

*Code 4.3. Covariance computation*

```
3.1  thrust::device_vector<float> d_cov(BANDS * BANDS);
3.2  alpha = 1.f; beta = 0.f;
3.3  cublasSgemm (handle,
        CUBLAS_OP_T, //the d_image_in is transposed to get column-major ordered format op(A)
        CUBLAS_OP_N,          //op(B)
        BANDS,    //rows of d_image_in (A) and d_cov (C)
        BANDS,    //columns of d_image_in (B) and d_cov (C)
        PIXELS,   //columns of d_image_in (A) and rows of d_image_in (B)
        &alpha,   //scalar set to~1
        thrust::raw_pointer_cast(d_image_in.data()), //matrix d_image_in (A)
        PIXELS,   //leading dimension of d_image_in (A)
        thrust::raw_pointer_cast(d_image_in.data()), //matrix d_image_in (B)
        PIXELS,   //leading dimension of d_image_in (B)
        &beta,    //scalar set to 0
        thrust::raw_pointer_cast(d_cov.data()),   //matrix d_image_in (C)
        BANDS    //leading dimension of d_image_in (C));

3.4  thrust::transform
        (d_cov.begin(),
        d_cov.end(),
        thrust::make_constant_iterator((float) (PIXELS - 1)),
        d_cov.begin(),
        thrust::divides<float>());
```

4. *Eigenvector decomposition*. This stage gets the eigenvectors associated to the covariance matrix computed in the previous stage by means of the cyclic Jacobi, which is divided into three parts:

   4.1. The *first part of this stage* deals with a sequential code on the host side, which is in charge of finding the maximum element of the covariance matrix that is not greater than a provided stop factor ($\varepsilon$). The calculation of the maximum value and its position ($row_{max}$, $col_{max}$) are carried out over the upper triangular matrix. After obtaining the maximum value position, the Equations (2)–(5) are calculated, where Equation (2) in this implementation is as the Equation (7), and *i* and *j* corresponds with the position of the maximum element in the covariance matrix ($row_{max}$, $col_{max}$). Moreover, the set of operations in (13) is computed over the identity matrix.

$$\begin{aligned} identity\ [row_{max}, row_{max}] &= cosA \\ identity\ [col_{max}, col_{max}] &= cosA \\ identity\ [row_{max}, col_{max}] &= sinA \\ identity\ [col_{max}, row_{max}] &= -sinA \end{aligned} \quad (13)$$

   where *l* = 1. *BANDS* (columns of the covariance matrix).

   4.2. The *second part of this stage* is executed on the GPU (device). A CUDA kernel is launched with the maximum number of threads per block and so many blocks as the relation between the number of *BANDS* and the $GPU_{BLOCKSIZE}$ (1024 threads):

$$\begin{aligned} threadsPerBlock &= GPU_{BLOCKSIZE} \\ numberBlocks &= \tfrac{BANDS+threadsPerBlock-1}{threadsPerBlock} \\ kernel &\ll \langle numberBlocks, threadsPerBlock \gg \rangle (\ldots) \end{aligned} \quad (14)$$

This kernel computes in parallel the Jacobi method with the equations in (15):

$$\begin{aligned}
d_{cov}[row_{max}, l] &= cosA \cdot d_{cov}[l, row_{max}] - sinA \cdot d_{cov}[l, col_{max}] \\
d_{cov}[col_{max}, l] &= cosA \cdot d_{cov}[l, col_{max}] + sinA \cdot d_{cov}[l, row_{max}] \\
d_{cov}[l, row_{max}] &= d_{cov}[row_{max}, l] \\
d_{cov}[l, col_{max}] &= d_{cov}[col_{max}, l]
\end{aligned} \quad (15)$$

As the goal of this stage is zeroing the off-diagonal elements, both parts of this stage are repeated until a stop condition is fulfilled. Because each iteration of this stage may undo the zeroes that are obtained in previous iterations, these iterations are repeated until the maximum value of the covariance matrix becomes smaller than a provided stop factor ($\varepsilon$). For each iteration, the covariance matrix becomes a diagonal matrix whose main diagonal represents the eigenvalues. Moreover, the eigenvector matrix is computed in parallel on the host side, as the equation shown in (16) using the cuBLAS library.

$$P = identity_1 \cdot identity_2 \cdot identity_3 \cdot \cdots \cdot identity_k \quad (16)$$

where *k* represents the last iteration of the eigenvector decomposition step.

4.3. The *last part of this stage* consists of extracting and sorting both eigenvalues and eigenvectors. The eigenvalues are located in the main diagonal of the covariance matrix obtained in parallel in the last stage. A kernel is in charge of extracting these eigenvalues in parallel from the covariance matrix (see line 4.3.2 of Code 4.4.3). This kernel is launched with one block of *BANDS* threads and each thread extracts its value of the main diagonal of the covariance matrix and stores it in a Thrust device vector (see lines 4.3.1.), *d_diagonal*.

*Code 4.4.3. Extraction and sort of eigenvalues and eigenvectors*

```
4.3.1   thrust::device_vector<float> d_diagonal(BANDS);
4.3.2   diagonalMain<<<1,BANDS>>> ( ... , ... ); //parameters covariance matrix and
        d_diagonal
4.3.3   thrust::device_vector<int> d_pos_data(BANDS*BANDS);
4.3.4   thrust::sequence(thrust::device, d_pos_data.begin(),
        d_pos_data.end(), 0);
4.3.5   thrust::sort_by_key
        (d_diagonal.begin(),
        d_diagonal.end(),
        d_pos_data.begin(),
        thrust::greater<float>());

4.3.6   sortEigenvectors <<<1,BANDS>>>
            ( ... , ... , ... ) //parameters: eigenvector matrix, the order of
        eigenvalues and the result of the sort
```

The main diagonal is sorted in descending order by means of two Thrust transformations: *thrust::sequence* and *thrust::sort_by_key*. The first transformation, *thrust::sequence*, fills a range with a sequence of numbers using the *thrust::device* execution policy for parallelization; the second transformation, *thrust::sort_by_key*, performs a key value sort, where the key value is provided by the *thrust::sequence* transformation. The result is a sorted device vector, *d_diagonal*, in descending order while using the *thrust::greater* comparison operator.

Moreover, the eigenvectors must be sorted according to the order that is followed by the eigenvalues. As each eigenvector contains *BANDS* elements, the sort operation is carried out changing the order of each eigenvector in matrix *P*, according to the order followed for the eigenvalues, which is kept in the *d_pos_data* device vector after sorting the eigenvalues. This sort is computed by a kernel, *sortEigenvectors*, which is launched with a block with *BANDS* threads. Each thread is in charge of an eigenvector.

5. *Projection and reduction*. In this stage, the original hyperspectral image is projected onto the first principal component, that is, onto the first eigenvector. Although the original hyperspectral image must be projected onto the set of eigenvectors, for simplicity this projection is performed only onto the first eigenvector. The projection is computed by multiplying the original hyperspectral image by a vector, the first eigenvector stored in matrix *P*. This operation is performed by a call to the *cublasSgemm* function (see Code 4.5).

*Code 4.5. Projection and reduction*

```
thrust::device_vector<float> d_projection (PIXELS,0);
thrust::device_vector<float> d_eigenvector (BANDS);
//copy the first eigenvector into d_eigenvector:
thrust::copy_n (thrust::device, d_G_ordered.begin(), BANDS, d_eigenvector.begin());
alpha = 1.0f; beta = 0.0f; //scalars for cublasSgemv
//d_projection = d_image_in x d_eigenvector (the first eigenvector):
cublasSgemm(handle,   //handle for use~cuBLAS
            CUBLAS_OP_N,
            CUBLAS_OP_N,
            PIXELS, BANDS,1,
            &alpha, //α scalar,
            thrust::raw_pointer_cast(d_image_in.data()),
            thrust::raw_pointer_cast(d_eigenvector.data()),
            &beta,   //β scalar
            thrust::raw_pointer_cast(d_projection.data()));
```

## 5. MPPA Implementation of PCA Jacobi

This section describes the MPPA-256-N parallel implementation of the PCA Jacobi Algorithm that is presented in Algorithm 1. Following the same structure as in Section 4.2, this section is divided into the four stages of PCA Jacobi Algorithm: image preprocessing, covariance computation, eigenvector decomposition, and projection and reduction.

1. *Image preprocessing*. This stage centers the image by computing and removing the average of each band of the image. Consequently, this computation is divided into two steps: firstly, the calculation of the average of each band of the hyperspectral image, and, secondly, the subtraction of this average from the original hyperspectral image.

    1.1 *Average per band calculation*. For each band of the original hyperspectral image (*BANDS* × *PIXELS* matrix), its average is calculated individually. This means that this step is band-wise parallelizable, and several averages are computed simultaneously. As described before, the main limitation of the MPPA-256-N is the cluster memory; consequently, to avoid additional iterations, each cluster needs enough memory to store, at least, one band of the image. For the images under study, this means a minimum of 0.2 MB and a maximum of 1.2 MB per band. Therefore, 16 bands can be mean-centered at the same time when exploiting all of the computational resources of the platform. As demonstrated in previous works [26], the communication between I/Os and clusters becomes the main bottleneck of this algorithm when processing very large datasets, such as hyperspectral images. As a result, this implementation takes advantage of all the available I/O resources

with the objective of reducing this bottleneck. As described before, there are two available I/Os, and each of them contains four DMAs to communicate with the clusters; hence, a total of eight communication links are available to send information to the clusters. The transmission of the bands to each cluster can be then parallelized, as depicted in Figure 5. As this implementation exploits the two available I/Os of the platform, in order to implement efficient communications, it has been decided that each I/O will handle the communications with half of the active clusters. As a result, I/O 0 communicates with clusters 0 to 7, while I/O does the same with clusters 8 to 15, and both of them exploit the four available DMAs. In addition, as each I/O has its own memory, there are two options to return information to the I/Os: (1) each cluster returns a value to its correspondent I/O and then I/O 0 and 1 exchange their partial information, or (2) each cluster sends the information to both I/Os simultaneously. The first option is a more computationally expensive operation, so throughout the implementation, the selected methodology has been the second one. Hence, in each iteration, the eight communication links are used in parallel to send one band to each cluster. Once the band has been received, the 16 internal cores—handled as threads—calculate its average by computing each one of them their share, as shown in lines 1.1–1.2 of Code 5.1. Afterwards, the master thread—i.e., the one supervising all the process—adds them up at the end, and orders the threads to start the average subtraction, as described in step 1.2.

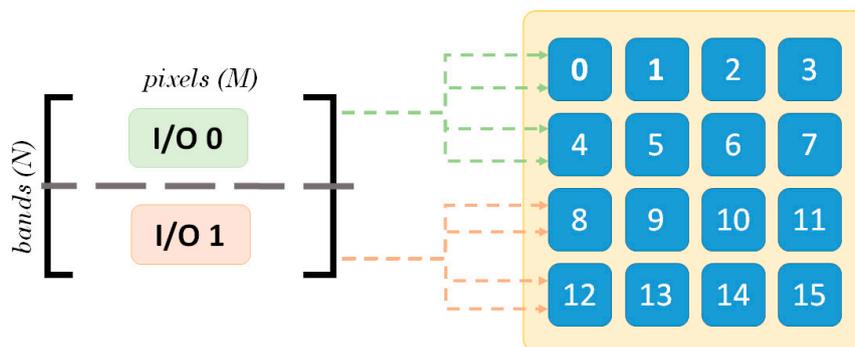

**Figure 5.** Division of the transmissions workload for the image preprocessing step.

1.2 *Subtraction of the average for each band*. Once the average has been computed, the internal threads subtract it to their share of the band, as depicted in lines 1.4–1.6 of Code 5.1. After that, the master thread returns the band, now mean-centered. As mentioned before, each cluster returns the information twice, one for each active I/O. Consequently, after this operation, each I/O contains a copy of the mean-centered image, which is necessary to start the covariance matrix computation.

*Code 5.1. Average calculation and removal*

```
1.1    for (i = thread_start; i < thread_stop; i++)
1.2        sum = sum + image_block[i];
1.3    Threads synchronization and average computation
1.4    average = sum / PIXELS;
1.5    for (i = thread_start; i < thread_stop; i++)
1.6        image_block[i] = image_block[i] - average;
```

2. *Covariance computation*. This stage calculates the covariance matrix associated to the original image after being mean-centered. As exposed in previous works, this stage becomes the main bottleneck of the algorithm in platforms that are memory bounded, as the MPPA-256-N [23].

Consequently, it is crucial to find an efficient implementation to reduce this bottleneck. In this case, the division of the workload of each I/O is such that each of them deals with half the active clusters. Likewise, each cluster will return its partial result twice, once per each active I/O, hence eliminating the communication between I/Os.

To compute the covariance matrix, the matrix that is obtained at stage 1 needs to be multiplied by its transpose. As noted in the previous step, depending on the input images, there are cases in which even one band of the image completely fills the cluster memory; therefore, neither of these matrices fit into a cluster, so the computation needs to be deeply iterated. Since each cluster can only store one band at a time, even computing just one element of the resulting matrix—i.e., multiplying two bands—must be an iterative process. As a result, the process of computing each element of the resulting matrix, as depicted in Figure 6, is divided following Equation (17):

$$E = (A \cdot C) + (B \cdot D) \tag{17}$$

As shown in Figure 6, $E$ represents the covariance matrix, $A$ and $B$ are the two halves of the input matrix, and $C$ and $D$ contain the two halves of its transpose, so each of them gathers $N$ bands with half the number of pixels, where $N$ represents the number of spectral bands of the original image. Consequently, $A$ and $B$ are $N \times M/2$ matrices, while $C$ and $D$ are $M/2 \times N$ matrices.

The process followed to compute the covariance matrix has been to calculate $A \times C$ in the clusters that are associated to I/O 0, while computing $B \times D$ in the clusters that are associated to I/O 1, and then add them up in each I/O. Furthermore, as the covariance matrix is symmetric, only half the computations are performed. It should be noticed that, for simplification, the process explained divides the computation in two halves. Nevertheless, when the size of the image grows, this implementation can be adapted to be divided in multiples of 2.

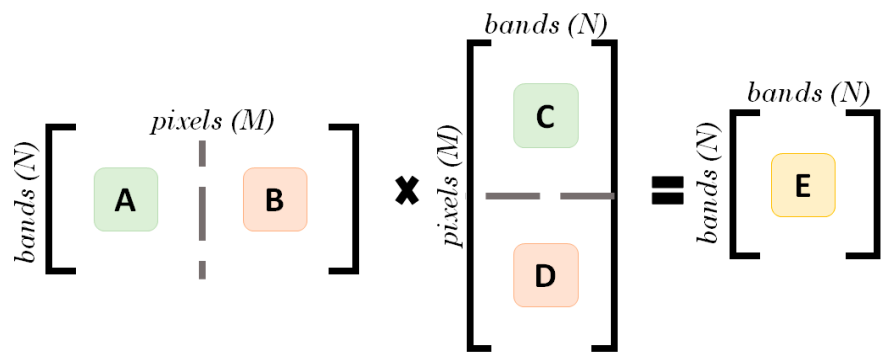

**Figure 6.** Division of the transmissions workload for the covariance computation step.

Finally, Code 5.2 provides the pseudocode for the operations performed in each one of the 256 cores. As expected, the operations performed within the code are quite simple, which highlights the communication-intensive nature of this step.

*Code 5.2. Covariance computation*

```
2.1    for (i = thread_start; i < thread_stop; i++)
2.2       partial_mult[thread_id] = partial_mult[thread_id] + A_block[i]*B_block[i]
```

3. *Eigenvector decomposition.* This stage computes the eigenvectors that are associated to the covariance matrix that were obtained in the previous stage. The algorithm selected to perform this calculation is the cyclic Jacobi described in Algorithm 1. This algorithm is mainly divided into two different processes: firstly, to find the elements of the matrix that have not been zeroed yet (see Code 5.3.1); secondly, to zero the elements selected in the first place, updating the matrix containing the eigenvectors (see Codes 5.3.2 and 5.3.3). Related with the architecture under study, the covariance matrix is rather small, so it fits into a cluster. As a result, it has been decided to implement the Jacobi algorithm using only one cluster and its internal 16 cores, since using all of the clusters introduces a communication overhead that would outpace any speedup provided by the exploitation of the rest of the clusters. Hence, the 16 internal cores are the ones that are responsible for parallelizing this algorithm. Specifically, the cyclic Jacobi method has been implemented, as follows:

*Code 5.3.1. Search elements to zero*

```
3.1.1     cond1 = |element| > ε;
3.1.2     cond2 = element not zeroed yet; (row-wise order)
3.1.3     cond3 = row, column not already used; (parallel Jacobi condition)
3.1.4     while (! End of Jacobi)
3.1.4        for (i = 0; i < BANDS-1; i++)
3.1.5           for (j = i+1; j < BANDS; j++)
3.1.6              if (cond1 && cond2 && cond3)
3.1.7                 Selection of aᵢⱼ, aᵢᵢ, aⱼⱼ
3.1.8                 Computation of α, cosα, senα (for building matrix P)
3.1.9                 Storage of i, j for the selected element
3.1.10                values_iter++;
3.1.11       if (values_iter == 0)
3.1.12          if (all elements selectable)
3.1.13             End of Jacobi
3.1.14          else
3.1.15             Reset cond2
3.1.16       else
3.1.17          Reset cond3
3.1.18          Update rows (all threads)
3.1.19          Update columns (all threads)
```

- The master thread—namely, the core executing the *main* function—handles the search of the next elements to be zeroed. In each iteration, this thread verifies that the stop condition has not been fulfilled yet and it then selects the maximum number of elements that can be simultaneously zeroed in the next iteration. As described in Algorithm 1, two elements can be zeroed in parallel as long as they do not share an index, i.e., that they are not located in the same row or column. Once all of these elements have been selected, this thread builds the rotation matrix and sends it to the processing cores. This process is iterated until the stop condition is fulfilled by all the off-diagonal elements, which is when the master thread sorts the eigenvalues in a descending order, together with their associated eigenvectors.
- Similarly, the cores perform the operation shown in (8), where *k* represents the current iteration, $P_k$ is the rotation matrix associated to the *k*th iteration, $C_{k-1}$ contains the covariance matrix modified in the *k*-1th iteration and $C_k$ is the resulting matrix, i.e., the covariance matrix with more elements already zeroed. To simplify this process, these products are not computed entirely, since the *P* matrix is mainly composed of zeros. As a result, these products are reduced to updating the rows and columns of *C*. The rationale behind this is that, as only the products that are related to the non-zero positions of *P* are computed, first the contribution of the products to the rows of *C* is computed, and afterwards, the same process is carried out for the columns, as depicted in Codes 5.3.2 and 5.3.3, respectively. Additionally, when updating the columns (Code 5.3.3), matrix *P* is also updated following Equation (10). The methodology to compute these products is very similar to the one that is described in (8), as each element to be zeroed only modifies four elements of its associated $P$ matrix.

*Code 5.3.2. Update rows*

```
3.2.1    for each element processed by each thread
3.2.2        for (k = 0; k < BANDS; k++)
3.2.3            new_a_ik = p_ii * a_ik + p_ji * a_jk;
3.2.4            new_a_jk = p_ij * a_ik + p_jj * a_jk;
3.2.5            a_ik = new_a_ik;
3.2.6            a_jk = new_a_jk;
```

*Code 5.3.3. Update columns*

```
3.3.1    for each element processed by each thread
3.3.2        for (k = 0; k < BANDS; k++)
3.3.3            new_a_ki = a_ki * p_ii + a_kj * p_ji;
3.3.4            new_a_kj = a_ki * p_ij + a_kj * p_jj;
3.3.5            a_ki = new_a_ki;
3.3.6            a_kj = new_a_kj;
3.3.7            new_p_ki = p_ki * p_ii + p_kj * p_ji;
3.3.8            new_p_kj = p_ki * p_ij + p_kj * p_jj;
3.3.9            p_ki = new_p_ki;
3.3.10           p_kj = new_p_kj;
```

4. *Projection and reduction*. This stage projects the original hyperspectral image onto the set of eigenvectors computed in stage 3, obtaining the principal components in the columns of the resulting matrix. However, for the images that are under study, the volume of spectral information retained in the first principal component is higher than an 80% on average, so this step has been simplified to just projecting the image onto the first eigenvector. In that way, the complexity of this stage is considerably reduced, as instead of multiplying two matrices, it only needs to multiply the original image by the first eigenvector. Related with the parallelization process, the method that is applied is similar to that of stage 2, but is much simpler. As one eigenvector fits into a cluster, it is broadcasted to all of them, while the original image is split in a pixel-wise order and is sent iteratively, where each core computes its share of the projection, as shown in Code 5.4. As happened in stages 1 and 2, each I/O manages one half of the image and sends it to its corresponding clusters. Afterwards, each cluster returns its partial results to each I/O.

*Code 5.4. Projection and reduction*

```
4.1    for (i = thread_start; i < thread_stop; i++)
4.2        for (iter = 0; iter < BANDS_PCA; iter++)
4.3            proj = 0;
4.4            for (j = 0; j < BANDS; j++)
4.5                proj = proj + image[i*BANDS+j] * eigenvecs[iter+BANDS_PCA+i];
4.6                pca_out [i*BANDS_PCA+iter] = proj;
```

## 6. Materials

The performance of the proposed implementations of the Jacobi-based PCA algorithm has been evaluated with synthetically generated hyperspectral images as well as with different real hyperspectral images.

The synthetic images have been generated with data extracted from the United States Geological Survey (USGS) database [27], which provides a library of spectral signatures from hundreds of different materials. The objective of using these images is to study the effect of the variation of the image dimensions on the performance that is given by both implementations. In this sense, two different sets of images have been created: in the first one the spatial resolution is varied while the spectral one remains unchanged (100 × 100 × 50, 200 × 200 × 50, 300 × 300 × 50, 400 × 400 × 50 and 500 × 500 × 50), whereas in the second one, the opposite procedure has been carried out (300 × 300 × 20, 300 × 300 × 30, 300 × 300 × 100 and 300 × 300 × 200). Specifically, these images have been generated by combining 10 different signatures that were randomly extracted from the USGS database and an additive noise drawn from a Gaussian distribution with an SNR set to 70 dB. Figure 7 shows properly scaled pseudo-color versions of the synthetic images that were generated with different spatial size.

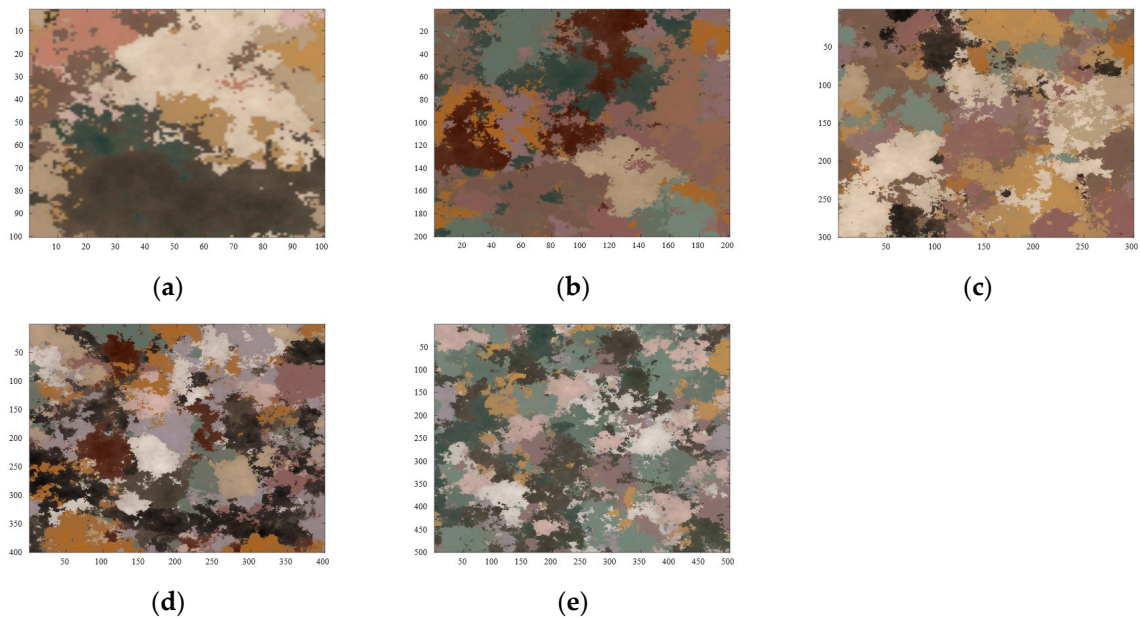

**Figure 7.** Pseudocolor version of the 100 × 100 × 50 (**a**); 200 × 200 × 50 (**b**); 300 × 300 × 50 (**c**); 400 × 400 × 50 (**d**); and 500 × 500 × 50 (**e**) synthetically generated images.

The real images utilized in this work were collected by the Airborne Visible/Infrared Imaging Spectrometer (AVIRIS). This sensor captures 224 spectral bands in the wavelength range of 0.4 to 2.5 μm. The first image used was taken over the region of *Cuprite*, Nevada, in 1997. The selected scenes consist of 250 × 191 pixels and 350 × 350 pixels. Several bands have been removed due to water absorption and low SNR, resulting in a total of 188 spectral bands. The site is well understood mineralogically, and it has several exposed minerals of interest, including alunite, buddingtonite, calcite, kaolinite and muscovite. The second hyperspectral image used in this work was recorded by AVIRIS over the standard scenes of *Jasper Ridge* Biological Preserve in California (1989). Water absorption and low SNR bands were removed prior to the analysis. The full data consists of a total amount of 512 × 614 pixels. The last hyperspectral image used was captured over the *World Trade Center* (WTC) area in New York City on 16 September 2001, five days after the terrorist attacks that collapsed the two main towers and other buildings in the WTC complex. The full data consists of a total amount of 614 × 512 pixels. Table 1 summarizes the features of the real hyperspectral images considered in this work while Figure 8 shows the pseudo-color versions of these images.

Table 1. Hyperspectral images for the experiments.

|  | Pixel Number | Bands |
| --- | --- | --- |
| Cuprite (small) | 47,750 | 188 |
| Cuprite (large) | 122,500 | 188 |
| Jasper | 314,368 | 224 |
| World Trade Center (WTC) | 314,368 | 224 |

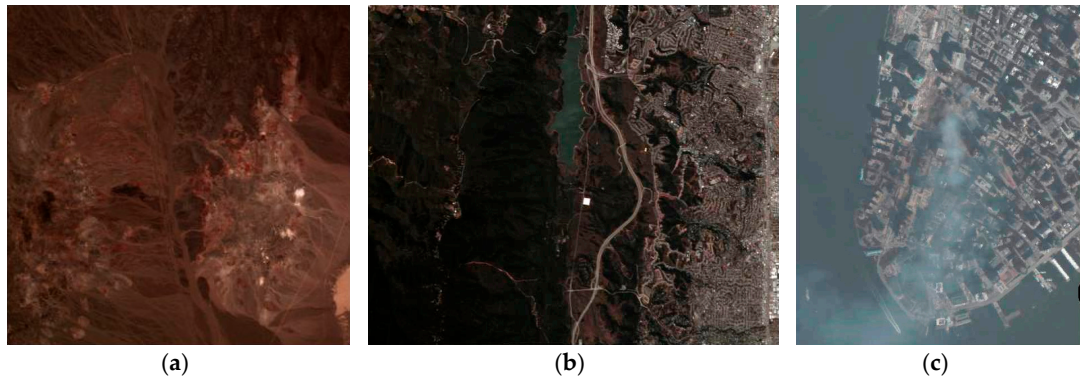

| (a) | (b) | (c) |

**Figure 8.** Pseudocolor version of the Cuprite large (**a**), Jasper Ridge (**b**) and World Trade Center (WTC) (**c**) Airborne Visible/Infrared Imaging Spectrometer (AVIRIS) images.

These two sets of images (synthetic and real) represent a heterogeneous test bench, according to the different spatial and spectral resolution of the images as well as different content, which allows for studying the behavior of the proposed implementations of the PCA algorithm under different situations.

## 7. Experimental Results

The experimental results have been measured for the hyperspectral images that are mentioned in the previous section and two different architectures, a NVIDIA GPU and a manycore platform from Kalray. In this section, the results that were achieved by both platforms are presented. Furthermore, to provide a comparison with a sequential implementation, the results have also been measured in a CPU.

### 7.1. Performance of the Sequential PCA Jacobi Algorithm

Before assessing the performance of the implementations that are described in Sections 4 and 5, first the algorithm has been evaluated executing a C implementation on an Intel® Core™ i7-4790 CPU with 32GB of external memory and four 3.6 GHz cores running an Ubuntu 14.04 LTS distribution. As the objective of this test is to serve as a sequential version, the implementation has been run using one core and one thread of the processor. Table 2 presents the results measured when executing the algorithm without the optimization flag (–O3) for obtaining the first principal component with the AVIRIS images.

**Table 2.** Execution times of the sequential Principal Component Analysis (PCA) Jacobi Algorithm using AVIRIS images (ms).

|  | Total |
| --- | --- |
| Cuprite (small) | 3672.6 |
| Cuprite (large) | 9410.2 |
| Jasper | 33,858.2 |
| WTC | 33,858.0 |

*7.2. Performance of the CUDA PCA Jacobi Algorithm*

The CUDA PCA Jacobi has been run in the NVIDIA GeForce GTX 680 GPU, which has a Kepler architecture. Table 3 summarizes the main features of this GPU.

**Table 3.** GPU Features.

|  | **GeForce GTX 680** |
|---|---|
| GPU cores | 1536 |
| Total amount of global memory (MB) | 2047 |
| GPU maximum clock rate (MHz) | 1058 |
| Maximum Graphics Card Power (W) | 195 |
| Minimum System Power Requirement (W) | 550 |
| Memory clock rate (MHz) | 3004 |

Table 4 provides the execution times of the CUDA PCA Jacobi algorithm for the aforementioned synthetic hyperspectral images. In this table, the processing times for obtaining 1, 3, and 5 principal components are also provided to analyze the effect of obtaining more than one component. The results in Table 4 show that the eigenvector decomposition is, by far, the most time consuming stage. This is due to the first part of this stage, which is calculated sequentially on the host side. The second most time consuming stage is the preprocessing due to the fact that it includes the transfer from the host to the device in order to load the initial hyperspectral image onto the global memory of the device. These transfers are carried out over the PCI express bus, taking a considerable amount of time, but thanks to this strategy, the execution time of the rest of the stages is much lower because they work over data stored on the device, which avoids the need of new transfers from the host to the device. Furthermore, the results that are provided in Table 4 prove that, in this implementation, increasing the spectral resolution of the image has a larger impact on the performance than augmenting the spatial one.

**Table 4.** Execution times for the Compute Unified Device Architecture (CUDA) PCA Jacobi Algorithm using synthetic images (ms).

|  | **Stage 1: Image Prep.** | **Stage 2: Cov. Matrix** | **Stage 3: Eigenvector Decomp.** | **Stage 4: Projection and Reduction** | | **TOTAL** |
|---|---|---|---|---|---|---|
|  |  |  |  | **#PCs** | **Time** |  |
| $100 \times 100 \times 50$ | 0.317 | 0.2177 | 68.21 | 1 | 0.453 | 69.19 |
|  |  |  |  | 3 | 0.541 | 69.28 |
|  |  |  |  | 5 | 0.559 | 69.30 |
| $200 \times 200 \times 50$ | 0.886 | 0.234 | 70.47 | 1 | 1.557 | 73.14 |
|  |  |  |  | 3 | 1.566 | 73.15 |
|  |  |  |  | 5 | 1.589 | 73.17 |
| $300 \times 300 \times 50$ | 1.560 | 0.286 | 70.91 | 1 | 2.981 | 75.73 |
|  |  |  |  | 3 | 2.987 | 75.74 |
|  |  |  |  | 5 | 2.993 | 75.74 |
| $400 \times 400 \times 50$ | 2.296 | 0.236 | 73.34 | 1 | 4.858 | 80.73 |
|  |  |  |  | 3 | 4.846 | 80.71 |
|  |  |  |  | 5 | 4.853 | 80.72 |
| $500 \times 500 \times 50$ | 3.467 | 0.983 | 77.53 | 1 | 7.277 | 89.25 |
|  |  |  |  | 3 | 7.290 | 89.27 |
|  |  |  |  | 5 | 7.406 | 89.38 |
| $300 \times 300 \times 20$ | 1.465 | 0.223 | 67.82 | 1 | 1.357 | 70.86 |
|  |  |  |  | 3 | 1.363 | 70.87 |
|  |  |  |  | 5 | 1.466 | 70.97 |
| $300 \times 300 \times 30$ | 1.461 | 0.197 | 66.66 | 1 | 1.888 | 70.20 |
|  |  |  |  | 3 | 1.905 | 70.22 |
|  |  |  |  | 5 | 1.909 | 70.22 |
| $300 \times 300 \times 100$ | 1.482 | 0.244 | 80.37 | 1 | 5.585 | 87.68 |
|  |  |  |  | 3 | 5.593 | 87.68 |
|  |  |  |  | 5 | 5.612 | 87.65 |
| $300 \times 300 \times 200$ | 1.605 | 1.332 | 146.32 | 1 | 11.23 | 160.48 |
|  |  |  |  | 3 | 11.12 | 160.37 |
|  |  |  |  | 5 | 11.02 | 160.35 |

Certainly, this table shows that, when increasing 25 times the spatial resolution of the image (i.e., comparing 100 × 100 × 50 and 500 × 500 × 50), the processing time is approximately multiplied by 1.3. However, when the spectral resolution is increased 10 times, the processing time is multiplied by an approximate factor of 2.25. This agrees with the fact that the eigenvector decomposition of the covariance matrix (which is a BANDS × BANDS matrix) represents the main bottleneck of this implementation.

Table 5 outlines the results that were obtained when processing the real hyperspectral images mentioned in the previous section, where the data in brackets in the rightmost column of the table represents the speedup factor that is achieved with respect to the sequential implementation. From this table, we can extract the same conclusions than with the previous one about the critical stage of the design (eigenvector decomposition) and about the greater impact that has an increase in the spectral resolution than in the spatial resolution of the images under processing.

**Table 5.** Execution times for the CUDA PCA Jacobi Algorithm using AVIRIS images (ms).

|  | Stage 1: Image Prep. | Stage 2: Cov. Matrix | Stage 3: Eigenvector Decomp. | Stage 4: Projection and Reduction | | TOTAL |
|---|---|---|---|---|---|---|
|  |  |  |  | #PCs | Time |  |
| Cuprite (small) | 1.220 | 0.982 | 140.65 | 1 | 5.834 | 148.68 (24.70) |
|  |  |  |  | 3 | 5.834 | 148.68 |
|  |  |  |  | 5 | 5.852 | 148.70 |
| Cuprite (large) | 2.289 | 1.374 | 148.90 | 1 | 13.92 | 166.48 (56.52) |
|  |  |  |  | 3 | 14.00 | 166.56 |
|  |  |  |  | 5 | 14.00 | 166.56 |
| Jasper | 4.638 | 1.043 | 170.57 | 1 | 40.53 | 216.78 (156.62) |
|  |  |  |  | 3 | 41.93 | 218.18 |
|  |  |  |  | 5 | 41.92 | 218.17 |
| WTC | 4.690 | 1.263 | 169.47 | 1 | 40.54 | 215.96 (156.77) |
|  |  |  |  | 3 | 40.78 | 216.20 |
|  |  |  |  | 5 | 40.63 | 216.05 |

*7.3. Performance of the MPPA PCA Jacobi Algorithm*

The PCA Jacobi has also been run in the MPPA-256-N architecture described before. Table 6 summarizes the main features of this platform, which is composed of 16 clusters with 16 cores each.

**Table 6.** MPPA-256-N Features.

|  | MPPA-256-N |
|---|---|
| Processing cores | 256 |
| Total amount of global memory (MB) | 32 |
| MPPA clock rate (MHz) | 400 |
| Typical power consumption (W) [28] | 10–20 |

Likewise, Tables 7 and 8 show the execution time of the PCA Jacobi algorithm running on the MPPA-256-N manycore architecture for the hyperspectral images that are presented before. In addition, the processing times for obtaining 1, 3, and 5 principal components are also provided to analyze the effect of computing more than one component.

Table 7. Execution times for the MPPA PCA Jacobi Algorithm using synthetic images (ms).

|  | Stage 1: Image Prep. | Stage 2: Cov. Matrix | Stage 3: Eigenvector Decomp. | Stage 4: Projection and Reduction | | TOTAL |
|---|---|---|---|---|---|---|
|  |  |  |  | #PCs | Time |  |
| 100 × 100 × 50 | 8.5 | 112.6 | 25.0 | 1 | 1.6 | 140.1 |
|  |  |  |  | 3 | 1.7 | 140.4 |
|  |  |  |  | 5 | 1.8 | 140.5 |
| 200 × 200 × 50 | 12.5 | 113.9 | 32.8 | 1 | 4.8 | 152.2 |
|  |  |  |  | 3 | 5.0 | 152.7 |
|  |  |  |  | 5 | 5.3 | 153.2 |
| 300 × 300 × 50 | 25.4 | 129.9 | 28.9 | 1 | 10.0 | 169.7 |
|  |  |  |  | 3 | 10.5 | 170.7 |
|  |  |  |  | 5 | 11.1 | 171.7 |
| 400 × 400 × 50 | 38.3 | 162.2 | 28.7 | 1 | 17.9 | 209.9 |
|  |  |  |  | 3 | 18.9 | 211.7 |
|  |  |  |  | 5 | 19.8 | 213.9 |
| 500 × 500 × 50 | 55.8 | 229.1 | 25.0 | 1 | 27.5 | 282.5 |
|  |  |  |  | 3 | 29.0 | 285.2 |
|  |  |  |  | 5 | 30.6 | 288.2 |
| 300 × 300 × 20 | 11.6 | 28.3 | 7.9 | 1 | 9.3 | 46.4 |
|  |  |  |  | 3 | 9.6 | 47.2 |
|  |  |  |  | 5 | 9.9 | 47.8 |
| 300 × 300 × 30 | 15.0 | 54.0 | 14.4 | 1 | 9.5 | 78.8 |
|  |  |  |  | 3 | 10.1 | 80.1 |
|  |  |  |  | 5 | 10.3 | 80.5 |
| 300 × 300 × 100 | 43.1 | 461.0 | 122.5 | 1 | 11.0 | 595.4 |
|  |  |  |  | 3 | 12.0 | 596.5 |
|  |  |  |  | 5 | 12.9 | 597.8 |
| 300 × 300 × 200 | 76.0 | 1,730.3 | 841.5 | 1 | 13.9 | 2586.6 |
|  |  |  |  | 3 | 16.0 | 2587.5 |
|  |  |  |  | 5 | 19.9 | 2591.2 |

Table 8. Execution times for the MPPA PCA Jacobi Algorithm using AVIRIS images (ms).

|  | Stage 1: Image Prep. | Stage 2: Cov. Matrix | Stage 3: Eigenvector Decomp. | Stage 4: Projection and Reduction | | TOTAL |
|---|---|---|---|---|---|---|
|  |  |  |  | #PCs | Time |  |
| Cuprite (small) | 42.8 | 1465.7 | 515.1 | 1 | 7.7 | 1989.5 (1.84) |
|  |  |  |  | 3 | 8.8 | 1990.5 |
|  |  |  |  | 5 | 10.6 | 1992.4 |
| Cuprite (large) | 103.7 | 1696.6 | 479.1 | 1 | 18.0 | 2194.6 (4.28) |
|  |  |  |  | 3 | 20.4 | 2199.1 |
|  |  |  |  | 5 | 25.3 | 2204.1 |
| Jasper | 293.7 | 4141.6 | 749.1 | 1 | 52.2 | 4944.2 (6.84) |
|  |  |  |  | 3 | 60.2 | 4959.2 |
|  |  |  |  | 5 | 84.8 | 4982.9 |
| WTC | 284.1 | 4144.2 | 770.9 | 1 | 52.7 | 4969.1 (6.81) |
|  |  |  |  | 3 | 59.4 | 4980.3 |
|  |  |  |  | 5 | 84.5 | 5004.4 |

The results provided in Tables 7 and 8 show that, as expected, the main bottleneck of the system is still the covariance matrix computation, regardless of the image used as input. Effectively, clusters memories have the capacity for only one image band. In addition, there are only 16 clusters at the MPPA-256-N. Consequently, since memory is a scarce resource at the cluster level, a costly iterative process is required. From these tables, it can also be inferred that the impact of increasing the number of principal components to compute is almost negligible. Additionally, these tables show that, for images with the same resolution but with very different contents—i.e., AVIRIS Jasper Ridge and WTC—similar processing times are obtained. This means that, although the PCA algorithm is data-dependent, the effect of the features (hyperspectral signatures) of the image under processing on the global execution time is almost negligible.

Furthermore, the results that are provided in Table 7 prove that, as it happens with the GPU implementation, in the MPPA implementation, increasing the spectral resolution of the image has a larger impact on the performance than augmenting the spatial one. This table shows that, when increasing 25 times the spatial resolution of the image (i.e., comparing 100 × 100 × 50 and 500 × 500 × 50), the processing time is approximately multiplied by 2. However, when the spectral resolution is increased 10 times, the processing time is multiplied by almost 56. This agrees with the covariance matrix being the main bottleneck, as described before, computing the covariance matrix involves multiplying the image by itself, generating as a result a *BANDS* × *BANDS* matrix. Hence, the greater the number of bands, the greater the number of elements that need to be computed in this step and the greater the processing time needed for this step, as shown in Table 7. Moreover, it is also coherent with Jacobi step behavior: as previously described, the Jacobi step is in charge of diagonalizing the covariance matrix; as a result, the complexity of this task will increase with the size of the covariance matrix, and thus with the number of bands, as shown in the execution times that were measured for this step.

*7.4. Comparisons*

When comparing these results with those that were obtained with the NVIDIA GPU, it is clear that the results obtained with the GPU for the images considered in this work are much faster than those that were obtained with the manycore architecture. In addition, it can also be observed that, for the smallest synthetic image in terms of spectral resolution (300 × 300 × 20), the MPPA is faster than the GPU. This demonstrates that, as expected, the memory is the main limitation of the MPPA. Effectively, for small images, the processing times can be even better than those of the GPU; however, when the memory limit is exceeded, the performance drops. Additionally, other features should be considered when comparing both implementations: for instance, GPU implementations have higher power consumption than MPPA. When huge amounts of data are just transmitted and then off-line processed on Earth, power dissipation does not represent a problem, being the preferred platform the one that is able to process the information in the shortest period of time. On the other hand, Earth Observation applications that demand on-board computing should have less power consumption and PCA Jacobi implementation on a NVIDIA GPU would not be the best choice indeed. In this sense, NVIDIA is doing huge research efforts in the consumption power field and last NVIDIA GPUs have improved it without decreasing their performance. For example, the maximum power dissipation for the NVIDIA GeForce GTX 1030, which follows the Pascal architecture, is 30 W and its recommended system power is 300 W.

In those cases in which power consumption becomes a tight constraint, architectures, like the MPPA, provide a fair trade-off between performance and energy efficiency. This architecture is designed for timing predictability and energy efficiency [29,30], achieving a typical power consumption of 10–20 W. As power efficiency requirements are gaining momentum, Kalray is also focusing its efforts towards this field. Specifically, the third generation of the MPPA processor will be launched soon, targeting critical embedded high-performance applications, and doubling the computing capabilities while maintaining the power consumption below 20 W.

As shown in Tables 3 and 6, the volume of resources of each piece of architecture is very different, so it should be taken into consideration when comparing both implementations. To do so, this paper proposes normalization with respect to the number of processing cores and the frequency to provide a fair comparison between the two implementations. This comparison is based on the number of hyperspectral cubes per second that each system can process. This metric—hereafter, *CPS*, *Cubes Per Second*—is obtained by computing the inverse of the total execution time of the algorithm for each platform, and provides a general idea of the throughput of the system. Nevertheless, this throughput is still highly dependent on the volume of resources of the platform; as a result, dividing this throughput by the number of processing cores and the operating frequency provides a fair metric to compare two implementations of the same algorithm that use a very different volume of resources. Consequently, this metric gives an idea of the efficiency of the implementation, independently of the available resources, as it provides the number of processed cubes per second and per core and MHz.

In this sense, Tables 9 and 10 gather the comparison of the two implementations for both the synthetic images and the AVIRIS ones, respectively. This table provides the number of hyperspectral cubes per second, *CPS*, when obtaining one principal component and the comparison parameter, measured in CPS per processing core and per MHz. As can be observed, the number of CPS is much higher for the GPU for both types of images. On the other hand, when considering the frequency and the number of cores, the MPPA proves to be more efficient (in terms of the figure of merit expressed in the rightmost column of Tables 9 and 10) when dealing with images with a low number of spectral bands (100 or less), being the efficiency of both implementations is very similar when processing images with a high number of spectral bands (AVIRIS images and the synthetic image with 200 spectral bands). Again, it is worth noting that, for smaller images, MPPA efficiency is considerably increased. This is coherent with what has been previously described, as the MPPA-256-N is memory bounded, and therefore, its throughput worsens when the memory usage is more intensive.

Table 9. Comparison of GPU and MPPA implementations of PCA Jacobi algorithm for synthetic images.

|  | Image | Time (ms) | Cores | Frequency (MHz) | CPS | CPS/(Core × MHz) |
|---|---|---|---|---|---|---|
| GPU | $100 \times 100 \times 50$ | 69.19 | 1536 | 1058 | 14.49 | $8.916 \times 10^{-6}$ |
|  | $200 \times 200 \times 50$ | 73.14 |  |  | 13.67 | $8.411 \times 10^{-6}$ |
|  | $300 \times 300 \times 50$ | 75.73 |  |  | 13.21 | $8.128 \times 10^{-6}$ |
|  | $400 \times 400 \times 50$ | 80.73 |  |  | 12.39 | $7.624 \times 10^{-6}$ |
|  | $500 \times 500 \times 50$ | 89.25 |  |  | 11.20 | $6.891 \times 10^{-6}$ |
|  | $300 \times 300 \times 20$ | 70.86 |  |  | 14.11 | $8.682 \times 10^{-6}$ |
|  | $300 \times 300 \times 30$ | 70.20 |  |  | 14.24 | $8.762 \times 10^{-6}$ |
|  | $300 \times 300 \times 100$ | 87.68 |  |  | 11.38 | $7.002 \times 10^{-6}$ |
|  | $300 \times 300 \times 200$ | 160.48 |  |  | 6.23 | $3.833 \times 10^{-6}$ |
| MPPA | $100 \times 100 \times 50$ | 140.1 | 256 | 400 | 7.14 | $69.705 \times 10^{-6}$ |
|  | $200 \times 200 \times 50$ | 152.2 |  |  | 6.57 | $64.163 \times 10^{-6}$ |
|  | $300 \times 300 \times 50$ | 169.7 |  |  | 5.89 | $57.546 \times 10^{-6}$ |
|  | $400 \times 400 \times 50$ | 209.9 |  |  | 4.76 | $46.525 \times 10^{-6}$ |
|  | $500 \times 500 \times 50$ | 282.5 |  |  | 3.54 | $34.569 \times 10^{-6}$ |
|  | $300 \times 300 \times 20$ | 46.4 |  |  | 21.55 | $210.466 \times 10^{-6}$ |
|  | $300 \times 300 \times 30$ | 78.8 |  |  | 12.69 | $123.929 \times 10^{-6}$ |
|  | $300 \times 300 \times 100$ | 595.4 |  |  | 1.68 | $16.402 \times 10^{-6}$ |
|  | $300 \times 300 \times 200$ | 2586.6 |  |  | 0.39 | $3.775 \times 10^{-6}$ |

Table 10. Comparison of GPU and MPPA implementations of PCA Jacobi algorithm for AVIRIS images.

|  | Image | Time (ms) | Cores | Frequency (MHz) | CPS | CPS/(Core × MHz) |
|---|---|---|---|---|---|---|
| GPU | Cuprite (small) | 148.68 | 1536 | 1058 | 6.72 | $4.135 \times 10^{-6}$ |
|  | Cuprite (large) | 166.48 |  |  | 6.01 | $3.698 \times 10^{-6}$ |
|  | Jasper | 216.78 |  |  | 4.61 | $2.836 \times 10^{-6}$ |
|  | WTC | 215.96 |  |  | 4.63 | $2.849 \times 10^{-6}$ |
| MPPA | Cuprite (small) | 1989.5 | 256 | 400 | 0.50 | $4.909 \times 10^{-6}$ |
|  | Cuprite (large) | 2194.6 |  |  | 0.46 | $4.450 \times 10^{-6}$ |
|  | Jasper | 4944.2 |  |  | 0.20 | $1.975 \times 10^{-6}$ |
|  | WTC | 4969.1 |  |  | 0.20 | $1.965 \times 10^{-6}$ |

Additionally, Figure 9 shows the first principal component that was obtained for the AVIRIS images with both implementations, which demonstrate their functional equivalence in terms of dimensionality reduction.

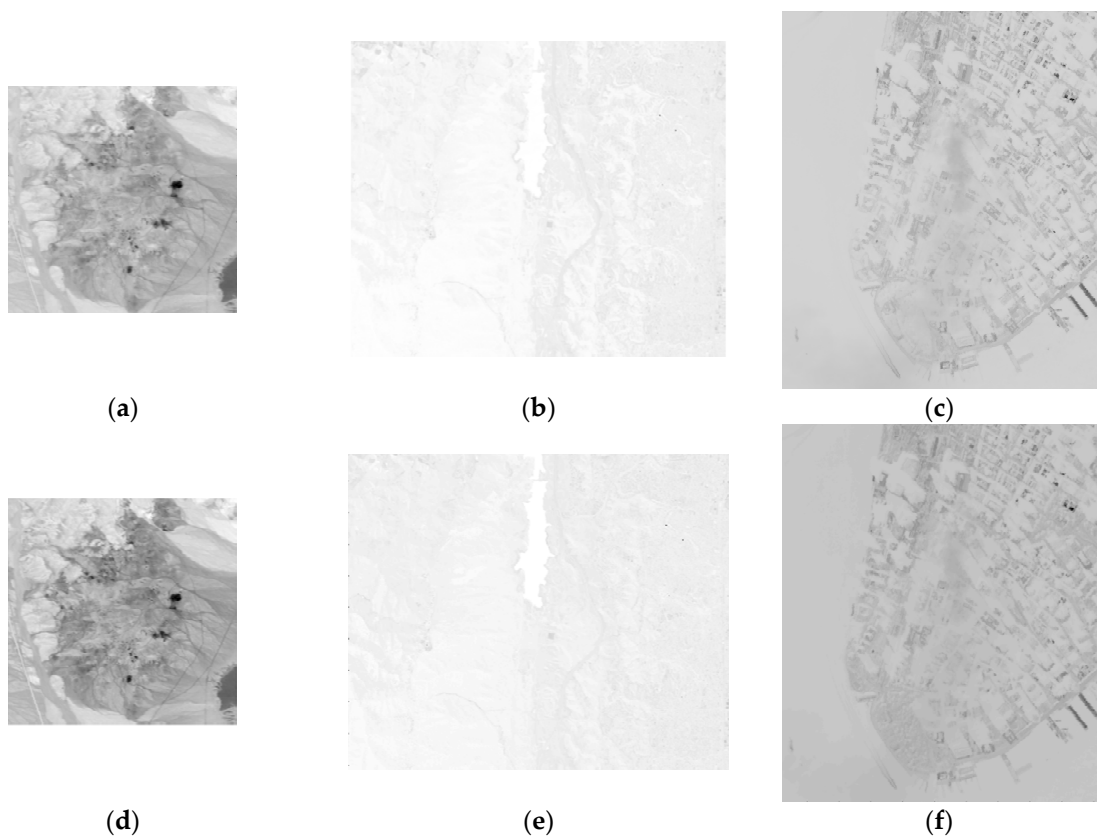

**Figure 9.** First principal component obtained with the GPU (**top row**) and with the MPPA (**bottom row**) for Cuprite (**a**,**d**); Jasper (**b**,**e**); and WTC (**c**,**f**) images.

After comparing the implementation of the PCA Jacobi algorithm, both on a GPU and on manycore architecture, a comparison with an implementation extracted from the state of the art of the same algorithm on a different platform has been also carried out in this work.

Specifically, in [10], Fernandez et al. present an implementation of PCA Jacobi on a Xilinx Virtex XC7VX690T FPGA, which is evaluated with two of the images used in this research work: AVIRIS Cuprite (large) and AVIRIS Jasper Ridge. The main characteristics of the FPGA targeted in the aforementioned work are summarized in Table 11.

**Table 11.** XC7VX690T Features.

|  | **XC7VX690T** |
| --- | --- |
| Logic Cells | 693,120 |
| DSP slices | 3600 |
| Intrenal memory (Kb) | 52,920 |
| Typical power consumption (W) [31] | Up to 60 |

To compare this implementation with the two ones presented here, a metric that is similar to the one described in the previous section will be applied. Since the criterion of the number of processing cores does not make sense in the FPGA case, only the operating frequency is included in the comparison metric. This comparison is presented in Table 12 and Figure 10, and it has been conducted using two AVIRIS images –Cuprite (large) and Jasper Ridge—as input, since they are the ones analyzed in [10].

**Table 12.** Comparison of GPU, MPPA, and field programmable gate array (FPGA) implementations of PCA Jacobi algorithm.

|  | Image | Time (ms) | Frequency (MHz) | CPS | CPS/MHz |
|---|---|---|---|---|---|
| GPU | Cuprite (large) | 166.48 | 1058 | 6.01 | $5.680 \times 10^{-3}$ |
|  | Jasper | 216.78 |  | 4.61 | $4.357 \times 10^{-3}$ |
| MPPA | Cuprite (large) | 2526.7 | 400 | 0.40 | $0.989 \times 10^{-3}$ |
|  | Jasper | 5488.8 |  | 0.18 | $0.455 \times 10^{-3}$ |
| FPGA | Cuprite (large) | 1490.0 | 76 | 0.67 | $8.831 \times 10^{-3}$ |
|  | Jasper | 4170.0 |  | 0.24 | $3.155 \times 10^{-3}$ |

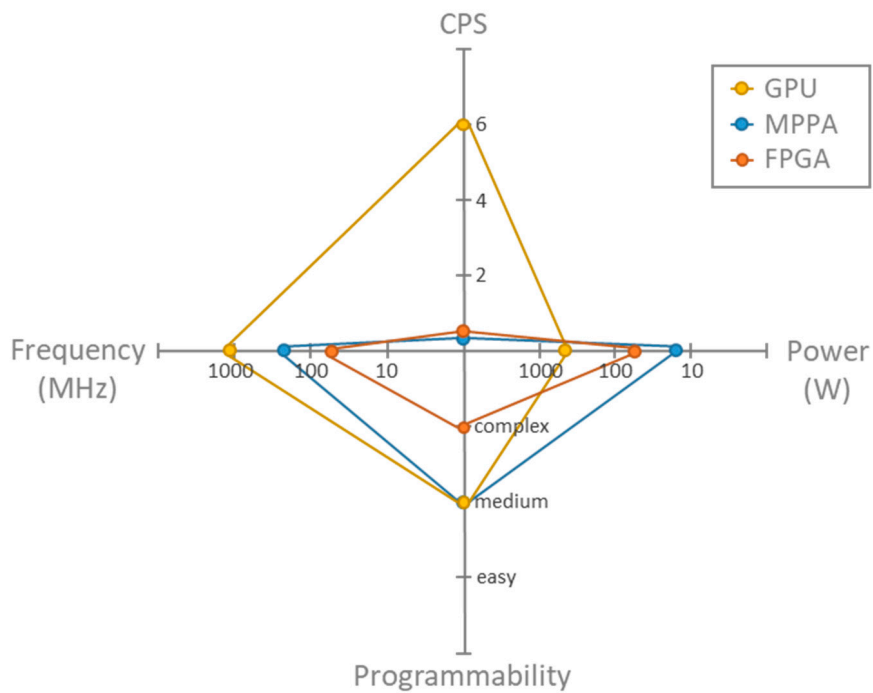

**Figure 10.** Comparison between platforms and implementations in terms of frequency, Cubes Per Second (CPS), power and programmability complexity.

Although the most efficient implementation is the FPGA one, as its operating frequency is smaller when compared with the other two (76 MHz), it is worth noting that this advantage is achieved at the expense of a smaller number of CPS when compared to a GPU and a design-time and power consumption increase (more complex programmability and up to 60 W of typical power consumption [31]) when compared to an MPPA. To delve into this triple comparison, Figure 8 presents a spider chart in which four key indicators are evaluated for each implementation: frequency, CPS (for simplicity purposes, only the values for the Cuprite image are included), power consumption, and programmability complexity. As can be observed, the GPU implementation stands out in terms of CPS; likewise, the MPPA one stands out in terms of energy efficiency. On the contrary, while the FPGA implementation stands out in terms of frequency, it is also the one that presents higher programmability complexity. Finally, Table 13 collects the processing times that were reported in papers [11,12], with similar images than the ones used in this work. Although these two works do not implement the Jacobi method, the processing times reported by their authors allow for checking the efficiency of our proposals in terms of the time that is required to process a given hyperspectral image.

**Table 13.** Processing times reported by other HPC (*High Performance Computing*) state-of-the-art works.

|  | Image | Time (s) |
| --- | --- | --- |
| GPU GTX 580 [11] | Cuprite (large) | 0.239 |
|  | WTC | 0.633 |
| Cloud computing * [12] | Cuprite (614 × 512 pixels × 224 bands) | 6.20 |

\* The cloud computing platform used for experimental evaluation in this work is built on a cluster that comprises nine nodes. The master node is the virtual machine, built on a host (IBM X3650M3) equipped with two Intel Xeon E5630 CPUs (eight cores in total) at 2.53 GHz, with 5 GB RAM and 292 GB SAS hard disk. The eight slave nodes are virtual machines built on 4 IBM BladeCenter HX5 blade computers. Each blade computer is equipped with two Intel Xeon E7-4807 CPUs (12 cores in total) at 1.86 GHz, and is connected to a 12 TB disk array by SAS bus.

## 8. Conclusions

In this paper, the implementation of the PCA algorithm onto a NVIDIA GPU and onto a Kalray manycore has been presented. In particular, the Jacobi method has been selected in both implementations in order to obtain the eigenvalue decomposition that is required by the PCA algorithm, as this method allows for exploiting the parallelism that is offered by both high-performance architectures. The procedure for accelerating the Jacobi-based PCA algorithm onto the two targeted devices has been exhaustively described and the main results that were obtained with a heterogeneous set of hyperspectral images have been reported. Moreover, a thorough analysis of the results that were achieved for each platform has been performed, introducing a novel comparison between them and with respect to an FPGA implementation recently published in the literature of the Jacobi-based PCA algorithm. Additionally, these results have also been compared with (1) similar algorithms in the recent literature, and (2) a sequential implementation of the algorithm. Last, but not least, the impact of the input image dimensions on the performance has been assessed by modifying both the spatial and the spectral resolution.

These comparisons have uncovered the pros and cons of each of the considered options in terms of the time that is required for processing a hyperspectral image, the power consumed for this processing, and the ease of programming of each platform, which provides crucial information for the potential designers of the next generation of hyperspectral imaging systems. Specifically, the results obtained prove that, although in general lines GPUs provide better performance rates, when features like power consumption are key decision factors, manycore architectures—such as MPPA—also provide efficient results. Its main limitation has also proven to be the internal memory, since for the smallest images it even outperforms the GPU under study. However, once this limit is exceeded, its performance drops substantially. As a result, the trade-off that is required for each application between performance, power consumption, and memory usage can help in deciding which is the most suitable platform.

**Author Contributions:** Ernestina Martel and Raúl Guerra programmed the PCA Jacobi algorithm on NVIDIA GPU; Raquel Lazcano and Daniel Madroñal programmed the PCA Jacobi algorithm on MPPA; José López and Sebastián López conceived and designed the experiments; Rubén Salvador and Eduardo Juárez performed the experiments; Cesar Sanz and Roberto Sarmiento supervised the technical work and several paper reviews. All authors contributed to the interpretation of the results and the writing of the paper.

**Acknowledgments:** This research has been partially funded by the European Commission through the ECSEL Joint Undertaking (ENABLE-S3 project, No. 692455) and the Ministry of Economy and Competitiveness (MINECO) of the Spanish Government through the ENABLE-S3 project, No. PCIN-2015-225, REBECCA (Resilient EmBedded Electronic systems for Controlling Cities under Atypical situations) project, No. TEC2014-58036-C4-4-R, and PLATINO (Distributed HW/SW Platform for Intelligent Processing of Heterogeneous Sensor Data in Large Open Areas Surveillance Applications) project, Nos. TEC2017-86722-C4-1-R and TEC2017-86722-C4-2-R.

**Conflicts of Interest:** The authors declare no conflicts of interest.